\documentclass[runningheads]{llncs}
\usepackage{graphicx}
\usepackage{amsmath}
\usepackage{amsfonts}
\usepackage{multirow}
\usepackage{float}
\usepackage{multirow}
\usepackage{color}
\usepackage{xcolor}
\usepackage[colorlinks]{hyperref}
\usepackage{subcaption}
\usepackage{threeparttable}
\usepackage{subcaption}
\begin{document}
%
\title{Unsupervised augmentation optimization for few-shot medical image segmentation} 
%
%
\author{Anonymous}
\author{Quan Quan\inst{1,2} \and 
Shang Zhao\inst{2} \and
Qingsong Yao\inst{1} \and
Heqin Zhu\inst{2} \and
S. Kevin Zhou\inst{1, 2}
}
\institute{\
Key Lab of Intelligent Information Processing of Chinese Academy of Sciences (CAS), Institute of Computing Technology, CAS, Beijing 100190, China\\
 \email{\{yaoqingsong19\}@mails.ucas.edu.cn} \email{\{quanquan, xiaoli,zhoushaohua\}@ict.ac.cn}  \and Medical Imaging, Robotics, and Analytic Computing Laboratory and Engineering (MIRACLE) 
School of Biomedical Engineering \& Suzhou Institute for Advanced Research, University of Science and Technology of China, Suzhou 215123, China}

\maketitle  
\begin{abstract}
    The augmentation parameters matter to few-shot semantic segmentation since they directly affect the training outcome by feeding the networks with varying perturbated samples.
    However, searching optimal augmentation parameters for few-shot segmentation models without annotations is a challenge that current methods fail to address.
    In this paper, we first propose a framework to determine the ``optimal'' parameters without human annotations by solving a distribution-matching problem between the intra-instance and intra-class similarity distribution, with
    the intra-instance similarity describing the similarity between the original sample of a particular anatomy and its augmented ones and the  
    intra-class similarity representing the similarity between the selected sample and the others in the same class.
    Extensive experiments demonstrate the superiority of our optimized augmentation in boosting few-shot segmentation models. We greatly improve the top competing method by 1.27\% and 1.11\% on Abd-MRI and Abd-CT datasets, respectively, and even achieve a significant improvement for SSL-ALP on the left kidney by 3.39\% on the Abd-CT dataset.
\keywords{Few-shot \and Medical Image Segmentation \and Data Augmentation}
\end{abstract}

\section{Introduction}
Neural networks have proven their superpower in many fields~\cite{zhou2017deep,ouyang2020self,yao2021one}, taking medical imaging analysis a big step forward. Nowadays, deep neural networks are applied in more medical applications or research~\cite{shi2017multimodal,zaidi2010pet}. 
Most of them are supervised by properly-curated datasets with correct annotations. 
Due to the high price of curating large datasets, 
few-shot learning has been proposed to encode the medical images into discriminative representations of an unseen class from only a few labeled examples (marked as support) to make predictions for unlabeled examples (marked as query) without the need for re-training the model~\cite{quan2022images,yao2021one,sung2018learning}. For few-shot segmentation, some models inject support into the network as guiding signals~\cite{zhang2019pyramid,siam2020weakly,zhang2019canet,roy2020squeeze}, or leverage meta-learning~\cite{tian2020differentiable,hendryx2019meta}. Other few-shot segmentation models, such as PANet~\cite{wang2019panet}, can be trained without annotations by forcing the identical local patches augmented with different image operations and then make predictions effectively by searching regions with the maximum similarity of features~\cite{siam2019amp,wang2019panet,ouyang2020self,wang2022few}. 
For medical imaging, SSL-ALP~\cite{ouyang2020self} finds that replacing cropped local patches with superpixels in the training process improves the performance of PANet. \cite{wang2022few} introduces an additional self-reference module to further improve SSL-ALP.

During training few-shot models, there always arises a crucial issue of manually adjusting the training hyperparameters, including optimizer~\cite{quan2021multi}, loss function~\cite{barron2019general,baik2021meta}, etc. Specifically, as detailed in section~\ref{sec:abla}, augmentation parameters form a crucial part, which highly affects the final performance of neural networks~\cite{cubuk2019autoaugment,hataya2020faster}. 
Many ideas spring up to automate the process, \textit{e.g.}, differential augmentation is an effective technique to auto-tune the augmentation parameters~\cite{eriba2019kornia,li2020dada,zheng2022deep}. However, existing methods are mainly designed for datasets with a bulk of annotations, because the optimization is powered by the reduction of the loss between predictions and ground truth of the labeled data~\cite{cubuk2019autoaugment,cubuk2020randaugment}. To the best of our knowledge, \textbf{there is no research about optimizing augmentation parameters without any annotation.}
It still remains unclear what happens when the number of available annotations reduces to a few shots. 

In this paper, we aim to bridge this gap and guide the augmentation optimization for a few-shot segmentation model, which is rarely researched. 
It brings two classical problems: ``\textbf{what} and \textbf{how} to guide the augmentation parameter optimization?'' 
(1) To address \textbf{what}, as previously mentioned, SSL-ALP~\cite{ouyang2020self} demonstrates the remarkable effect of superpixels for few-shot medical segmentation, which provides a rationale for us to employ superpixels as alternative annotations to guide the optimization.
(2) To address \textbf{how}, we believe that a good set of augmentation parameters should produce more realistic samples that preserve sufficient class information. 
Thus, we suggest a way of quantifying the class information by computing the similarities among samples belonging to the same class, termed ``\textbf{intra-class similarity distribution}''. Moreover, we also show that the similarities among the augmented samples generated from a single instance, which we term ``\textbf{intra-instance similarity distribution}'', should correspond with the intra-class similarity distribution. The parameters that achieve this correspondence are the ones we expect.

\begin{figure}[t]
    \centering
    \includegraphics[width=\linewidth]{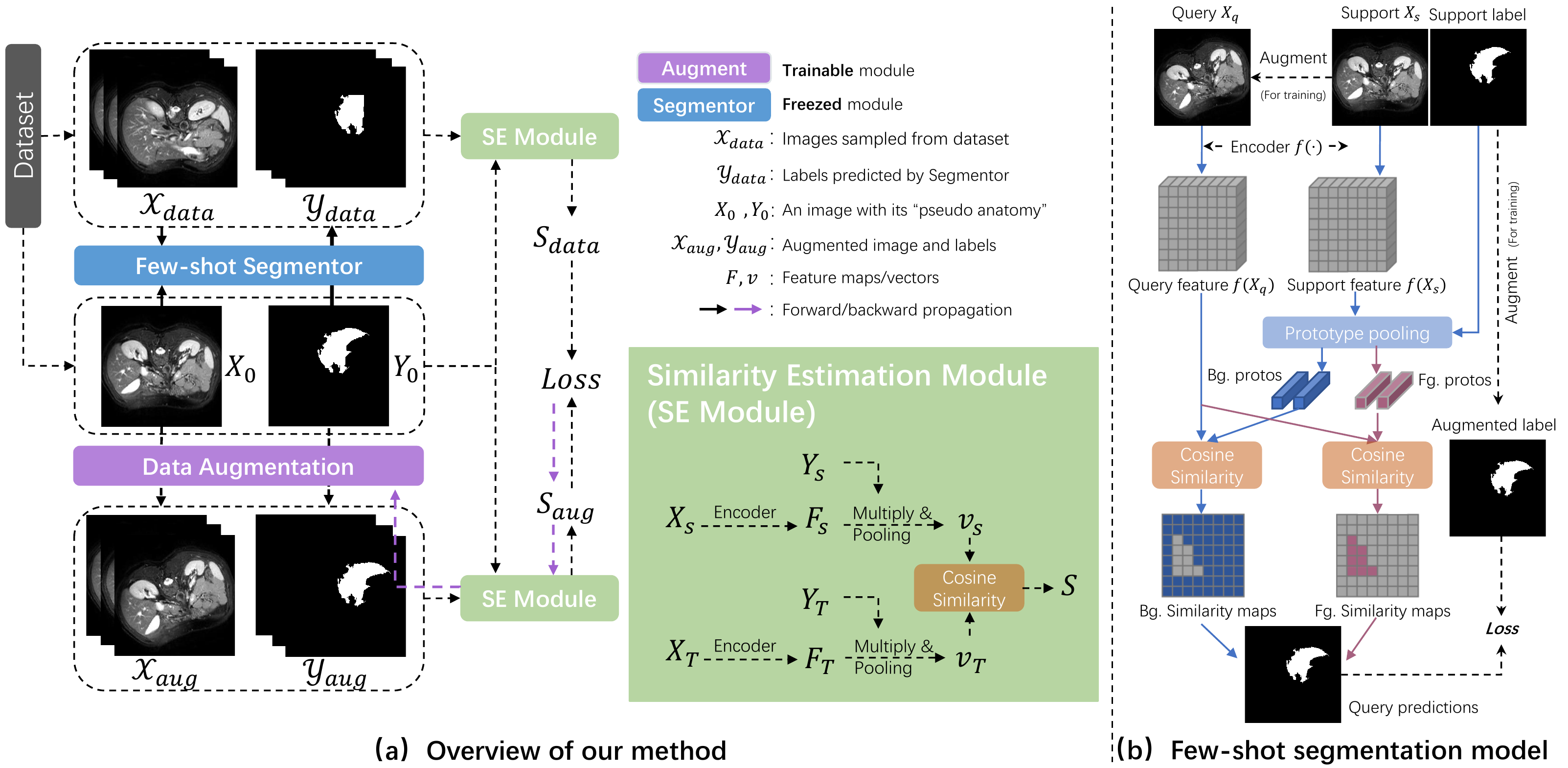}
    \caption{(a) Overview of our method and (b) pipeline of the few-shot segmentation models~\cite{ouyang2020self,wang2022few}.}
    \label{fig:overview}
\end{figure}
Based on the above, we propose a novel augmentation method that automatically tunes the parameters of the models by minimizing the two distributions. 
We leverage a pretrained segmentor with its built-in encoder to predict corresponding labels of superpixels to construct similarity distributions of real data; and apply differential augmentation to generate similarity distributions of augmented data. 
Our contributions are summarized as follows: 
(1) We \textbf{firstly} propose an augmentation parameter optimization framework for few-shot medical image segmentation task \textbf{without the help of any manual annotation};
(2) We conduct extensive experiments to prove the effectiveness of our proposal. The average Dice Similarity Coefficient (DSC) of the state-of-the-art model Self-ref~\cite{wang2022few} is significantly increased by \textbf{1.27\%} and \textbf{1.11\%} on Abd-MRI and Abd-CT datasets.

\section{Method}
\noindent{\textbf{The proposed method:}}
We mainly aim at improving the superpixel-based self-supervised segmentation methods \cite{ouyang2020self,wang2022few} trained without the requirement of any manual annotations (Figure~\ref{fig:overview} (b)). Specifically, we generate a superpixel-based pseudo label for each support image. We then apply geometric and intensity transformations on support image and mask to get the corresponding query image and mask (only transformed geometrically are applied on masks). Subsequently, the obtained support and query images and their associated masks are fed into our network for end-to-end training. For testing, a support image is fed to our trained model together with its manual annotations in order to predict the segmentation for a query image.

As in Fig.~\ref{fig:overview}, our main objective is to search the optimal augmentation parameters $\hat{A}$ (block in \textcolor{violet}{violet}) for a few-shot medical segmentation model $g(\cdot)$ based on a training dataset $D_{train}$ without any label. We accomplish this goal by aligning the \textbf{intra-instance similarity (IIS) distribution} $S_{aug}$ of augmented anatomies from a particular ``pseudo anatomy'' $Y \in \mathcal{Y} $ (a superpixel) with the \textbf{intra-class similarity (ICS) distribution} $S_{data}$ of the ``pseudo anatomy'' and its corresponding anatomies in other images via reducing $\mathcal{L}1$ distance. 

\begin{equation}
    \begin{aligned}
    \hat{A} = \arg\min_{A} \mathbb{E}_{Y \sim \mathcal{Y}} {\left\lVert \mu(S_{data}(Y)) - \mu(S_{aug}(Y, A))\right\rVert}_1.
    \end{aligned}
\end{equation}

To present our method clearly, we first introduce the details of multiple key ingredients in Section~\ref{sec:details} and then provide the justification in Section~\ref{sec:proof}.

\subsection{Key Ingredients} \label{sec:details}

\textbf{Intra-class similarity distribution $S_{data}$:} 
To obtain the ICS distribution, we first pre-train a segmentor $g(\cdot)$ with a built-in encoder $f(\cdot)$ following \cite{ouyang2020self}. 
Supposing the training dataset $\mathcal{X}_{train} = \{X_1, X_2, ..., X_n\}$ contains $n$ images with their superpixel labels $\mathcal{Y}=\{Y^1, Y^2, ...,Y^n\}$, we choose a image $X_i \in \mathbb{R}^{H \times W}$ from $\mathcal{X}_{train}$ as the source image with its $k$ superpixel annotations ${Y}^i \in \mathbb{R}^{K \times H \times W}$. We suppose one superpixel $Y^i_j \in \mathbb{R}^{H \times W}$ in ${Y}^i$ as an ``pseudo anatomy'' on $X_i$, and predict the ``pseudo anatomy'' $Y^i_j$ in other training images $\hat{\mathcal{Y}}^i_j$.
Finally, we estimate the similarities $S_{data} \in \mathbb{R}^B$ by comparing their features. 
\begin{equation}
\hat{Y}_k = g(f(X_i), Y^i_j, f(X_k)),
\end{equation}
\begin{equation}
    SE(X_i, Y^i_j, \mathcal{X}_{train}, \hat{\mathcal{Y}}^i_j) = \mathbb{E}_{ \{X_k, \hat{Y}_k\} \sim \mathcal{X}} \Big[ sim\Big(f(X_i) \otimes Y^i_j, f(X_k) \otimes \hat{Y}_k \Big) \Big], 
    \label{eq:se}
\end{equation}
\begin{equation}
    S_{data} = \Big\{  SE(X_i, Y^i_j, \mathcal{X}_{train}, \hat{\mathcal{Y}}^i_j) , X_i \sim \mathcal{X}_{train}  \Big\},
    \label{eq:s_data}
\end{equation}
where $SE(\cdot,\cdot)$ refers to \emph{similarity estimation module}; $\otimes$ contains multiplication following with average pooling; $sim(\cdot,\cdot)$ refers to cosine similarity.

\noindent{\textbf{Intra-instance similarity distribution $S_{aug}$:}}
In order to estimate the distribution of augmented images, we transform $X_i$ into $B$ different augmented images $\mathcal{X}^i_{aug}$ with different augmentation parameters sampled from $A$. 
In addition, augmented version $\mathcal{Y}^i_{aug}$ are also obtained with the same parameters.  
Similarly, as the above, we can easily estimate the cosine similarities $S_{aug} \in \mathbb{R}^B$ between all  $\mathcal{Y}^i_{aug}$ and $Y^i_j$,

\begin{equation}
    S_{aug} = \Big\{ SE(X_i, Y^i_j, \mathcal{X}^i_{aug},\mathcal{Y}^i_{aug}), X_i \sim \mathcal{X}_{train}  \Big\}.
    \label{eq:s_aug}
\end{equation}
Note that we sample the boundary parameters to maintain the marginal distribution with a sufficient quantity of class information so that the entire distribution preserves sufficient class information as well.

\noindent{\textbf{Loss function:}} 
Since the averages of similarity distribution $S_{data}$ and $S_{aug}$ dominates the performance, we apply a $\mathcal{L}1$ loss for minimizing the disparity of those averages in our method. 
Additionally, the less satisfied segmentor performance leads to a gap between the average of distribution $S_{data}$ and the ground truth distribution. Therefore, we introduce a correction parameter $\tau$ to ensure that the average $S_{aug}$ is always larger than the one of $S_{data}$. In practice, we set $\tau=0.03$.

\begin{equation}
    \mathcal{L} = {\left\lVert\mu(S_{data}) + \tau - \mu(S_{aug}) \right\rVert}_1.
    \label{eq:loss}
\end{equation}

\noindent{\textbf{Optimization:}} 
Instead of calculating $S_{data}$ with all training images, we evaluate $S_{data}$ by sampling images from batches to realize a memory-efficient scheme.
There are two steps to optimize the loss function (Eq.~\ref{eq:loss}): (1) Parameter initialization; (2) Gradient optimization. Through manual matching $S_{data}$ and $S_{aug}$, we can quickly identify a set of good parameters and customize the intensity of each kind of augmentation operation according to the practical application scenarios, but we still need a differential method to obtain precise parameters. 
Gradient optimization can perceive precise values but easily fall into a suboptimal local point. Therefore, we combine the two strategies: we leverage manual search for coarse-grained search and gradient optimization for fine-grained search. 
Additionally, we observe that $\mu(S_{aug})$ behaves more sensitive to the variations of color and rotation, and less sensitive to the scaling, and translation. 
Thus, the $l$ parameters are applied \emph{bundle optimization} to prevent $\mu(S_{aug})$ from being dominated by a few parameters.
\begin{equation}
    \hat{A} = \arg\min \mathcal{L}(A_0 + \beta \Gamma),
\end{equation}
where $A_0 \in \mathbb{R}^l$ and $\Gamma \in \mathbb{R}^l$ is initialized during parameter initialization.
\subsection{Justification} \label{sec:proof}
There are three key consensuses supporting our method: (1) More correct samples lead to a better model and more wrong samples lead to worse~\cite{sun2017revisiting}; (2) For an anatomy (an image patch), the encoded average feature vector $v \in \mathbb{R}^m$ with $m$ channels are concatenated from two vectors: $a \in \mathbb{R}^{k}$ containing only instance information with $m-k$ channels and $c \in \mathbb{R}^{m-k}$ containing only class information with $k$ channels. We denote the concatenation as $v = [a~c]$; (3) The augmentation transformations will keep class information as much as possible, that is, the augmentation operations will first discard the instance information and then class information.

Based on the above, our method is detailed in three parts: 
(1) Estimate the intra-class similarity distribution $S_{data}$; 
(2) Estimate the intra-instance similarity distribution $S_{aug}$;
(3) Match $S_{data}$ and $S_{aug}$ by reducing the $\mathcal{L}1$ distance between their average similarities $S_{data}$ and $S_{aug}$ to achieve the ``optimal'' augmentation parameters $\hat{A}$.

To obtain the class information $c$ of a particular class, we can estimate it from intra-class similarities by calculating similarities between different samples of the same anatomy. 
\begin{equation}
    \mu(S_{data}) = \frac{1}{n-1} \sum_i^{n-1} ([a_0~c]) ([a_i~c]) = a_0(\frac{1}{n-1}\sum_i^{n-1} a_i) + c^2 \approx c^2,
\end{equation}
where $\{a_i\}$ are supposed to be various vectors, so $\sum a_i$ can be seen as a very small value. 

Similarly, we can get features of augmented images $v' = [a'~c']$, where $a' \in \mathbb{R}^{k'}$ is the instance information and $c' \in \mathbb{R}^{m-k'}$ denotes the shared information between all $v'$. Thus, $\mu(S_{aug}) \approx c'^2$. Due to supposition (2), if $c'^2 \ge c^2$, $c'^2= [d~c]^2 = d^2 + c^2$, where $d$ is the remained instance information of $a_0$. If $c'^2 < c^2$ $c^2 = [d'~c']^2=d'^2+c'^2$, where $d'$ is missed class information.

We here suppose $N_r$ as the number of the correct samples keeping all the class information $c$ and $N_w$ for wrong samples not keeping all the class information. We suppose there is $b$ possible value in one channel, and we get
\begin{equation}
    \begin{aligned}
        &N_r = b^{k'}, N_w = 0, &when~ c' \ge c; \\
        &N_r = b^{k}, N_w = b^{k'} - b^k, &when~ c' < c.         
    \end{aligned}
\end{equation} 
Therefore, to achieve a more robust model by increasing $N_r$ and reducing $N_w$, we should make $c$ and $c'$ close. 





\section{Experiments}
\subsection{Experimental Setup}
\textbf{Datasets:} We perform evaluations under two datasets: Abd-CT and Abd-MRI. 
Specifically, 
\textbf{Abd-MRI}~\footnote{\url{https://chaos.grand-challenge.org/}} is from ISBI 2019 Combined Healthy Abdominal Organ Segmentation Challenge (Task 5) \cite{kavur2021chaos} containing 20 3D T2-SPIR MRI scans. 
\textbf{Abd-CT}~\footnote{\url{https://www.synapse.org/\#!Synapse:syn3193805/wiki/217789}} is from MICCAI 2015 Multi-Atlas Abdomen Labeling challenge \cite{landman2015miccai} containing 30 3D abdominal CT scans. 
We aim to segment four organs from both datasets, \textit{i.e.}, \textit{liver}, \textit{spleen}, \textit{left} and \textit{right kidneys}.
We follow \cite{ouyang2020self} to reformat 3D images as 2D slices and resize them to $256 \times 256$, and separate 2D slices into \textit{upper} and \textit{lower} abdomen groups to process \{\textit{liver}, \textit{spleen}\} and \{\textit{left}/\textit{right kidney}\}, respectively. For evaluation, we follow \cite{ouyang2020self} to perform experiments with five-fold cross-validation and quantified with Dice Similarity Coefficient (DSC).

\noindent{\textbf{Superpixel:}} Superpixels are generated by clustering local pixels using statistical models with respect to low-level image features. In this work, we follow \cite{ouyang2020self} to utilize an off-the-shelf and unsupervised graph-cut-based algorithm [68]. Additionally, to eliminate the superpixels belonging to the background, we ignore the superpixels with entropy lower than 3.

\noindent{\textbf{Differential augmentation operators:} }
To optimize augmentation parameters by gradients, we follow \cite{eriba2019kornia} to make the augmentation operators differentiable including gamma correction, rotation, translation, shearing and scaling. For uniform and Bernoulli distribution embedded in these operators, the backpropagation is implemented by score function estimator~\cite{schulman2015gradient}. Moreover, the transformation is implemented by \textit{pytorch autograd modules}.

\noindent{\textbf{Training details: }} All training and testing images are resized to $256 \times 256$. 
In the augmentation parameters optimization, the Adam optimizer is used with a learning rate of 0.0005. The maximum epoch is set at 400. The proposed method is implemented in PyTorch, and the training takes around 3 hours to finish on an Nvidia RTX.

\begin{table}[h]
    \centering
    \caption{Quantitative comparison between our method and other SOTA methods. The \textbf{best} and \underline{second-best} performances are highlighted.}
    \begin{threeparttable}
    \begin{tabular}{l|rrrr|r}
    
    \multicolumn{6}{c}{\multirow{1}{*}{\textbf{Abd-MRI} dataset}} \\
    \hline
    Method  & ~Liver & ~Spleen & ~Kidney(L) & ~Kidney(R) & ~Average \\
    \hline
    SE-Net~\cite{roy2020squeeze}*   & 27.43 & 51.80  & 62.11     & 61.32     & 50.66 \\
    PANet~\cite{wang2019panet}* & 42.26 & 50.90  & 53.45     & 38.64     & 46.33 \\
    GCN-DE~\cite{sun2022few}* & 49.47 & 60.63  & 76.07     & 83.03     & 67.30 \\
    \hline
    SSL-ALP~\cite{ouyang2020self}$^1$ & 72.94 & 67.02  & 73.63     & 78.39     & 72.99 \\
    SSL-ALP~\cite{ouyang2020self}$^1$~+~Ours & \underline{73.48} & \textbf{71.06} & \underline{76.24}  &  79.94    &  \underline{75.18} \\
    \hline
    Self-ref~\cite{wang2022few}$^2$ &  71.54 & 69.18  & 76.07     & \underline{80.45}     & 74.31 \\
    Self-ref~\cite{wang2022few}$^2$~+~Ours & \textbf{73.96} & \underline{69.79}  & \textbf{77.85}  & \textbf{80.73}     & \textbf{75.58} \\
    \hline
    \hline
    
    \multicolumn{6}{c}{\multirow{1}{*}{\textbf{Abd-CT} dataset}} \\
    \hline
    Method  & ~Liver & ~Spleen & ~Kidney(L) & ~Kidney(R) & ~Average \\
    \hline
    SE-Net~\cite{roy2020squeeze}*   & 0.27 & 0.23  & 32.83 & 14.34  & 11.91 \\
    PANet~\cite{wang2019panet}* & 38.42 & 29.59  & 32.34   & 17.37  & 29.43 \\
    GCN-DE~\cite{sun2022few}* & 46.77 & 56.53 & 68.13 & 75.50  & 61.73 \\
    \hline
    SSL-ALP~\cite{ouyang2020self}$^1$ & 73.40 & 63.82  & 63.68  & 56.52  & 64.35 \\
    SSL-ALP~\cite{ouyang2020self}$^1$~+~Ours~ & \underline{74.02} & 64.22  & \underline{67.06}  &  57.16 & 65.61 \\
    \hline
    Self-ref~\cite{wang2022few}$^2$ & \textbf{74.62} & \underline{65.95}  & 65.06   & \underline{61.45}   & \underline{66.77} \\
    Self-ref~\cite{wang2022few}$^2$~+~Ours & 73.70 & \textbf{67.64}  & \textbf{68.06}  & \textbf{62.12}  & \textbf{67.88} \\
    \hline
    \hline
    \end{tabular}
    \begin{tablenotes}
        \footnotesize
        \item[] * copied from~\cite{wang2022few}; $^1$ running official code; $^2$ personal implementation.
    \end{tablenotes}
    \end{threeparttable}
    \label{table:main}
\end{table}

\subsection{Results}
\begin{figure*}[h]
    \centering
    \begin{subfigure}{0.495\textwidth}
        \centering
        \includegraphics[width=0.9\linewidth]{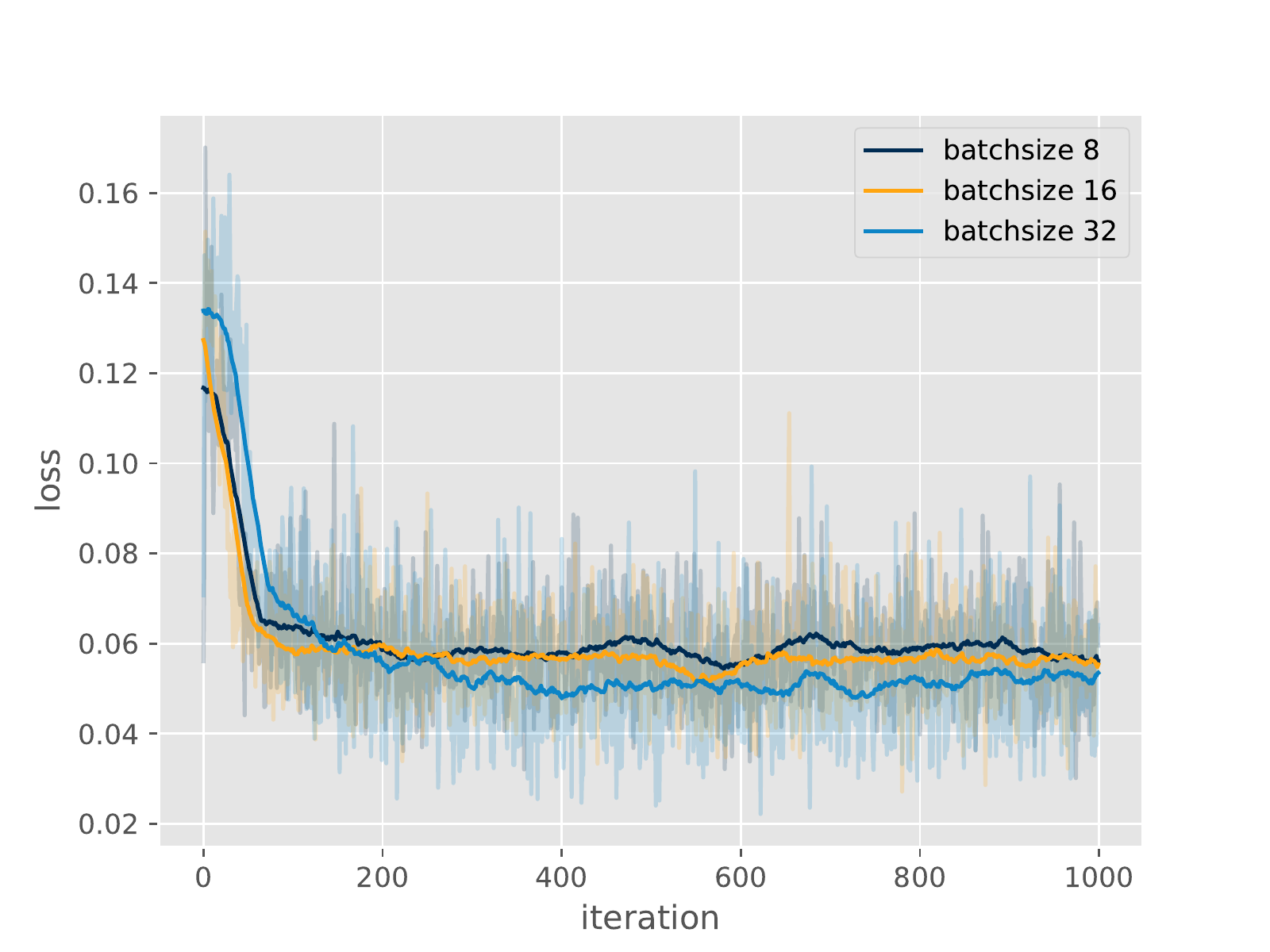}
        \caption{}
        \label{fig:bs}
    \end{subfigure}
    \begin{subfigure}{0.495\textwidth}
        \centering
        \includegraphics[width=0.9\linewidth]{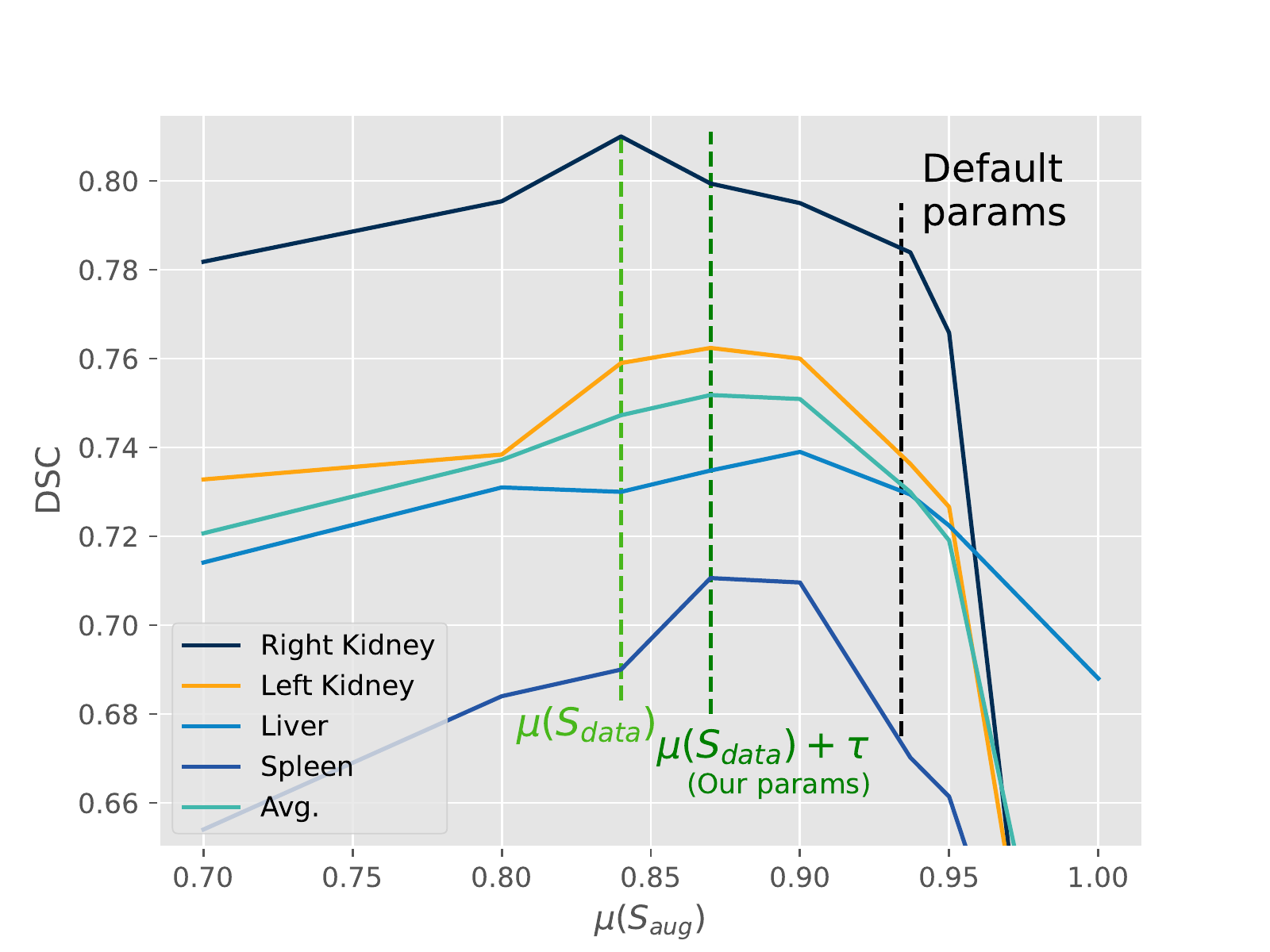}
        \caption{}
        \label{fig:s_aug}
    \end{subfigure}

    \begin{subfigure}{0.95\textwidth}
        \centering
        \includegraphics[width=1\linewidth]{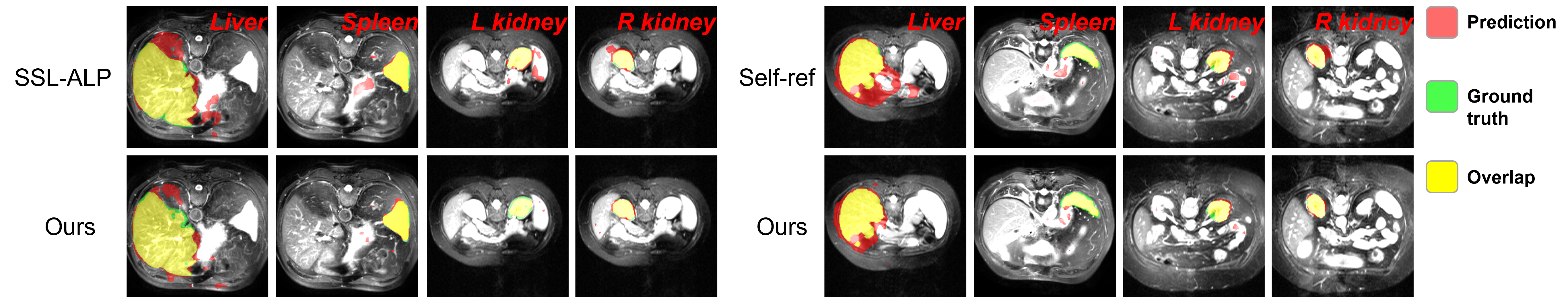}
        \caption{}
        \label{fig:2}
    \end{subfigure}
    \caption{Analysis experiments on Abd-MRI dataset. (a) Behaviors with different batch sizes; (b) DSC vs. $\mu(S_{aug})$; (c) Visual results on Abd-MRI dataset. }
\end{figure*}

We compare our method with popular approaches including PANet~\cite{wang2019panet}, SE-Net~\cite{roy2020squeeze}, GCN-DE~\cite{sun2022few}, SSL-ALPNet~\cite{ouyang2020self} and the SOTA Self-ref~\cite{wang2022few}. The results of PANet, SE-Net and GCN-DE are brought from \cite{wang2022few}. The SSL-ALP is run with official code, and Self-ref is our personal implementation. We use the same configuration for a fair comparison. 
As shown in Table~\ref{table:main}, our method boosts the models \cite{ouyang2020self,wang2022few} with impressive improvements. 
On the Abd-MRI dataset, our method achieves the best DSC of 75.58\%. The average DSC of SSL-ALP and Self-ref are increased by 2.16\% and 1.27\%, respectively. On the Abd-CT dataset, a similar performance improvement is observed. Specifically, our method achieves the best performance on almost every abdominal organ, except that the DSC of \textit{liver} in the last line drops but \textit{left kidney} takes a leap, which we think is caused by the variance of sensitivity to augment for different organs. 
Our visualizations in Fig.~\ref{fig:2} also demonstrate that our approach improves the quality of predictions.

\begin{table}[h]
    \centering
    \caption{Ablation study on SSL-ALP\cite{ouyang2020self} with Abd-MRI dataset}
    \begin{tabular}{l|rrrr|r}
        \hline
    Method & ~Liver & ~Spleen & ~Kidney(L) & ~Kidney(R) & ~Average\\
    \hline
    baseline  & 72.94 & 67.02  & 73.63     & 78.39     & 72.99 \\
    +Initialization  &73.10 &	68.22	& 76.06 & \textbf{80.18}  &	74.39 \\
    +Gradient optim ~  &	73.40&	70.52&	76.18&79.80	&	74.97 \\
    +Init +Gradient    & \textbf{73.48} & \textbf{71.06} & \textbf{76.24}  &  79.94    &  \textbf{75.18} \\
    \hline
    \hline
    w/o bundle optim  & 73.23 &	66.78	& \textbf{77.22} & \textbf{81.33}  &	74.63 \\
    w/ bundle optim  & \textbf{73.48} & \textbf{71.06} & 76.24  &  79.94    &  \textbf{75.18} \\
    \hline
    \end{tabular}
    \label{table:abla1}
\end{table}

\subsection{Ablation Study} \label{sec:abla}
\textbf{Initialization \& bundle optimization \& batchsize:}
During training, we sometimes find loss unstable at the start of training. Initialization by manually searching to match $\mu(S_{data})$ and $\mu(S_{aug})$ is helpful to ensure the parameters converge to the right place. 
As shown in Table~\ref{table:abla1}, manually minimizing the loss also works well. Afterward, we can leverage the differential operations to further improve the performance. 
Table~\ref{table:abla1} shows that bundle optimization can help improve the weak parts (\textit{liver} and \textit{spleen}).
Figure~\ref{fig:bs} displays that a large batch size benefits the training process, but it will cost more computation resources. Considering the GPU memories and training time, we set the batch size at 32.

\noindent{\textbf{Similarity distribution: }}
We manually control the augmentation parameters to train models with different $\mu(S_{aug})$. As shown in Figure~\ref{fig:s_aug}, when the $\mu(S_{aug})$ is slightly greater than $\mu(S_{data})$, the performance achieve the best. Additionally, the optimal augmentation parameters for each anatomy vary to some extent. For example, the optimal $\mu(S_{aug})$ for \textit{liver} is larger than \textit{right kidney}.

\section{Conclusion}
In this paper, we propose a novel unsupervised automatic augmentation framework for the few-shot medical segmentation models by aligning the intra-instance and intra-class similarity distribution. 
A number of few-shot medical segmentation models including the state-of-the-art model are significantly enhanced by our method according to our experiments.
The results of ablation studies also verify the effectiveness of our method from different perspectives.

\clearpage
\newpage
\bibliographystyle{splncs04}
\bibliography{reference}

\end{document}